\documentclass[lettersize,journal]{IEEEtran}
\usepackage{amsmath,amsfonts}
\usepackage{algorithmic}
\usepackage{algorithm}
\usepackage{array}
\usepackage[caption=false,font=normalsize,labelfont=sf,textfont=sf]{subfig}
\usepackage{textcomp}
\usepackage{stfloats}
\usepackage{url}
\usepackage{verbatim}
\usepackage{graphicx}
\usepackage{cite}
\usepackage{epsfig}
\usepackage{amssymb}
\usepackage{bbding}
\usepackage{booktabs}
\usepackage{array}
\usepackage{authblk}
\usepackage{multirow}
\usepackage{CJKutf8}           
\usepackage[utf8]{inputenc}
\begin{document}

\title{Yuan-TecSwin: A text conditioned Diffusion model with Swin-transformer blocks}

\author{Shaohua Wu, Tong Yu, Shenling Wang, Xudong Zhao}



\maketitle

\begin{abstract}
   Diffusion models have shown remarkable capacity in image synthesis based on their U-shaped architecture and convolutional neural networks (CNN) as basic blocks. The locality of the convolution operation in CNN may limit the model's ability to understand long-range semantic information. To address this issue, we propose Yuan-TecSwin, a text-conditioned diffusion model with Swin-transformer in this work. The Swin-transformer blocks take the place of CNN blocks in the encoder and decoder, to improve the non-local modeling ability in feature extraction and image restoration. The text-image alignment is improved with a well-chosen text encoder, effective utilization of text embedding, and careful design in the incorporation of text condition. Using an adapted time step to search in different diffusion stages, inference performance is further improved by 10\%. Yuan-TecSwin achieves the state-of-the-art FID score of 1.37 on ImageNet generation benchmark, without any additional models at different denoising stages. In a side-by-side comparison, we find it difficult for human interviewees to tell the model-generated images from the human-painted ones. 
\end{abstract}


\section{Introduction}
\IEEEPARstart{R}{ecently}, text-guided synthesis has become a universal method for image generation and edition\cite{saharia2022photorealistic}\cite{ramesh2022hierarchical}\cite{ramesh2021zero}. Based on large- scale multimodal datasets and scalability of the model, a series of generative models, including GAN, diffusion models, autoregressive model and flow model, have reached an unprecedented level on text- to-image synthesis and can even generate photorealistic images\cite{saharia2022photorealistic}\cite{ramesh2022hierarchical}\cite{feng2022ernie}\cite{balaji2022ediffi}.
Diffusion models benefit from the fixed learning process and the flexible model structure, and thus are widely applied in image synthesis. They are able to create images with high diversity without collapse, because of the ability to maintain semantic structure and fine-grained images. Compared to the diffusion models, Generative Adversarial Networks (GAN)\cite{brock2018large}\cite{kang2021rebooting} usually suffer unstable training during adversarial training and difficulties in generating diversified images. Autoregressive models \cite{yu2022scaling} and flow models often use an approximation as posterior probability and calculate in maximum likelihood. VAE have to use a surrogate loss, and the flow model relies more on a specific model structure. 

\begin{figure*}[t]
\centering
\includegraphics[width=1\textwidth]{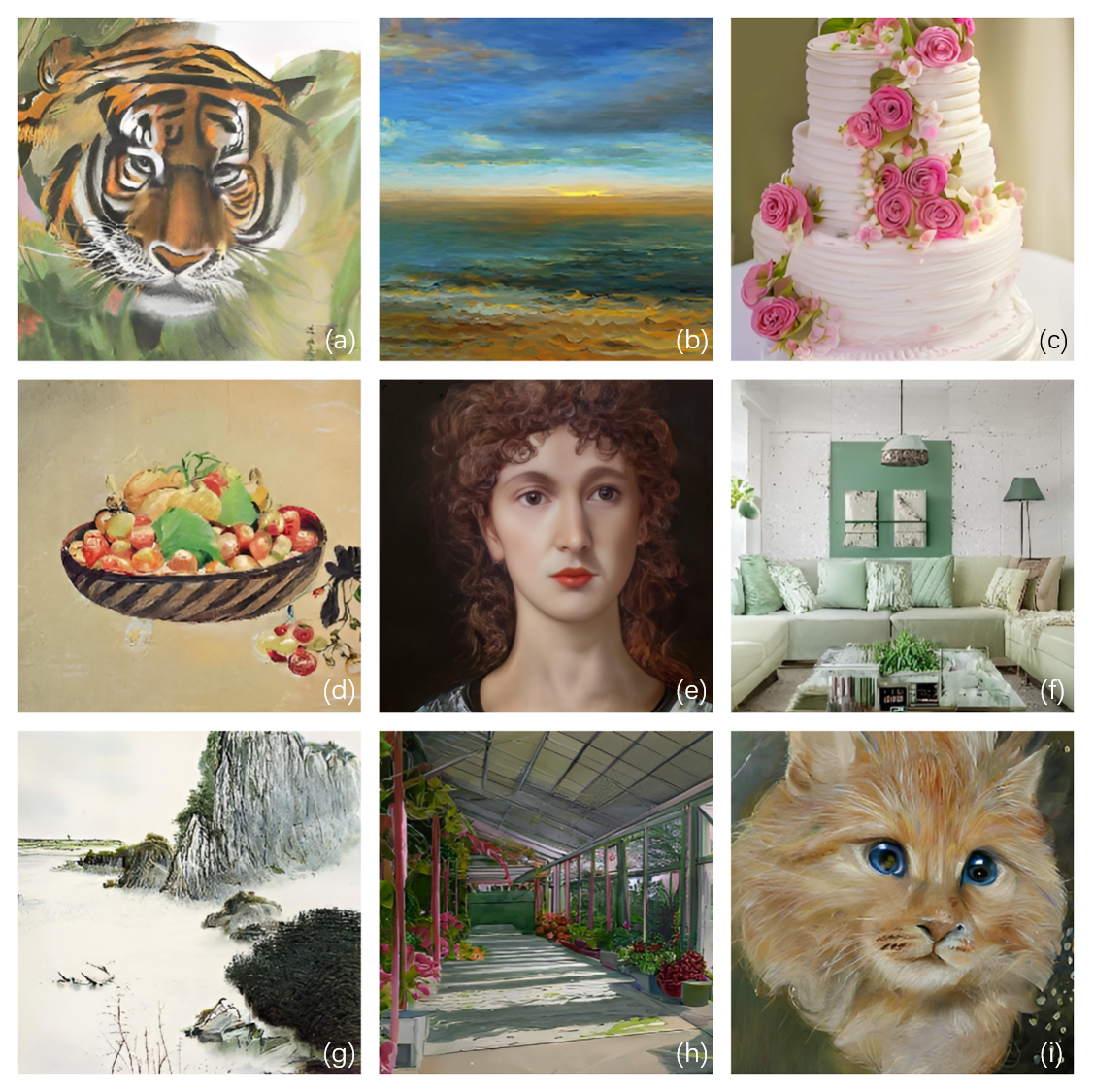}
\caption {Sample images generated by Yuan-TecSwin.(a) A painting in Chinese-style of a tiger couching in the bush (\begin{CJK}{UTF8}{gbsn}画一张国画，蹲在草丛中的老虎\end{CJK}); (b) Make a painting in Impressionism. Describing the sunrise over the sea. (\begin{CJK}{UTF8}{gbsn}画一幅印象派油画，描绘了海上日出的场景\end{CJK}); (c) A wedding cake with three layers, decorated with roses. (\begin{CJK}{UTF8}{gbsn}一个三层的婚礼蛋糕，蛋糕上装点着玫瑰花\end{CJK}); (d) A classic Chinese painting about a basket of fruits,  (\begin{CJK}{UTF8}{gbsn}一幅中国画，关于一篮子水果\end{CJK}); (e) An oil painting of a women with blonde curly hair (\begin{CJK}{UTF8}{gbsn}画一幅油画，一个女人，长着金色的卷发\end{CJK}); (f) This is an photo of living room with green wall. There are two couches in the living room with a tea table in front of them. There are plants on the tea table. (\begin{CJK}{UTF8}{gbsn}这是一张客厅的照片，客厅的墙是绿色的，摆着两个沙发，沙发前面有一个玻璃茶几，沙发上放着白色和绿色的靠枕，茶几上放着绿色植物\end{CJK}); (g) Make an ink and wash painting. Describing the mountains over clouds. (\begin{CJK}{UTF8}{gbsn}画一幅水墨画，描绘了云雾中的山峦\end{CJK}); (h)Corridor interior view of a greenhouse with many flowers planted and sunlight streaming into the greenhouse (\begin{CJK}{UTF8}{gbsn}一座温室的走廊内景，温室里种了很多花，有阳光照进温室中来\end{CJK}); (i) An oil painting of a golden kitty with blue eyes (\begin{CJK}{UTF8}{gbsn}画一幅油画，一只金色的猫咪，猫咪的眼睛是蓝色的\end{CJK}).}
\label{fig:toutu}
\end{figure*}

 Diffusion models with CNN-based blocks have achieved excellent result in image synthesis, with their capability in extracting low-level features such as the shape, color and texture of an object. However, in term of global or high-level visual information regarding to the relations and connections between objects, the CNN-based structure shows its weak points due to the capability in long-range semantic information. In a text-to-image generation task, it is important to regenerate basic features of an object, while the position, relative size, and depth of field are also significant elements to be considered. Motivated by the application of the transformer structure in NLP models, and the performance of Swin-transformer in computer vision domain, we propose a Swin-based Unet to improve the efficiency and balance the speed-performance trade-off in image generation. Compared to the words processed in NLP tasks, images, especially images with high resolution, has much higher resolution of pixels, which significantly increases the computational complexity. Swin-transformer combines the advantages of CNN and transformer, developed with a hierarchical feature map for a more computational-efficient adaption in the image domain. 

In addition to the model structure, the iterative sampling process from noise to meaningful images is also a key to the performance and inference time of the diffusion model. Nicole and Dhariwal\cite{nichol2021glide} find that the diffusion model can maintain high sample quality with fewer steps, which speeds up sampling in practical applications. We also notice a common setting in recent diffusion models that most of which usually consider global time steps as equal when calculating the loss function\cite{nichol2021glide}\cite{ho2020denoising}. In fact, in the denoising time steps when the image is slightly corrupted, the task of image recovery can only contribute less to the model training. In contrast, in the denoising time steps where the image is severely corrupted, the model will be forced to solve the tougher task only relying on a few identifiable entities\cite{choi2022perception}.

Open-source large-scale multimodel datasets lower the training threshold\cite{schuhmann2022laion}\cite{gu2022wukong}, but the presence of watermark, pornography, violence, and racial discrimination in datasets can also decrease the generation capability of the diffusion model and make it unsuitable for public release. To capture the characteristic of a large-scale dataset, the parameters of the main diffusion models reach billions to tens of billions, which increases training difficulty and computing resource consumption\cite{feng2022ernie}\cite{balaji2022ediffi}. In addition to data quality, the text encoder is another factor that influences the performance of text-to-image synthesis. It is common for a text-to-image model to employ a pre-trained language model to encode input texts into a sequence of embedding.

In this work, we propose Yuan-TecSwin, a text-to-image swin-based diffusion model dedicated to an eﬀicient learning of the denoising training process and a rational usage of dataset. The experiments show that the Swin-transformer based diffusion model achieves a performance similar to that of the CNN-based diffusion models, with only one-tenth of the parameters. Specifically, we build a Chinese multi-modal dataset with up to 360 million image-text pairs to ensure the diversity and the quality. The pre-trained base model is further finetuned with a human-labeled artistic dataset in higher quality to reinforce the ability in generating artworks. Instead of a uniform sampling during denoising time steps, the adapted time step searching is introduced to further strengthen the model for image recovery task. Figure \ref{fig:toutu} shows sample images generated by Yuan-TecSwin.

To summarize, the main contributions of our work are as follows:
\begin{enumerate}
\item{A novel text-conditioned diffusion model is constructed by introducing Swin-transformer blocks. The model we constructed can achieve a better image generation performance compared with a diffusion model with with similar architecture and CNN-blocks.}
\item{A versatile application of text encoder is discovered, that the average output of multiple layers brings a better result than the output of the final layer.}
\item{The sdapted time step searching method is proposed to refine the denoising process during diffusing steps, which allows a significant performance on image generation, without incorporating additional models.}
\item{A new state-of-the-art FID of 1.37 is achieved on ImageNet generation benchmark. We also design a test on human evaluation for image generation, and find it difficult for the interviewees to distinguish the generated pictures with the “real” one.}
\end{enumerate}

\section{Related Work}
\subsection{Text-conditional Diffusion Model.}
Diffusion Probabilistic models (DM) consisting of a forward process and a reverse process, have archived the new state-of-the-art results in varieties of generative tasks, especially in computer vision domain. Slow sampling is an inherent shortcoming for DM due to the use of Markov chain. Several studies has focused on eﬀicient  schedules to accelerate the sampling process, with partial sampling\cite{nichol2021improved}, or using differential equations for a learning-free sampling\cite{song2020score}. GENIE\cite{dockhorn2022genie} uses a second-order solver to optimize the sampling scheme. A more straightforward method is to truncate the diffusion and reverse process by early stopping\cite{lyu2022accelerating}. Knowledge distillation is also a feasible method to accelerate the sampling process. With this method, model only requires half diffusion steps compared to the original one\cite{salimans2022progressive}.
In our work, we follow the logic of DDPMs. The forward process is to gradually add Gaussian noise to an image until it becomes random noise to achieve the purpose of corruption. And the reverse process is a process of image noise reduction. it starts with a random noise, and the diffusion model gradually removes the noise until an image is generated.

\subsection{Other Generative Model for Image Synthesis}

Generative models learn the joint distribution among a certain data distribution and make a balance between versatility and computability. In addition to diffusion models, the family of deep generative models also include variational auto-encoder (VAE), flow model, generative adversarial networks and energy-based model. Before the prosperity of diffusion model, VAE, GAN and their derivative VQ-VAE, VQ-GAN are the dominance among the generative models in a long time. 
GAN-based models allow for flexible model structure and efficient sampling for images with high resolution\cite{brock2018large}\cite{kang2021rebooting}. However, the computation of likelihood are difficult to control, resulting in unstable training performance and difficulty in optimization. VAE-based methods also achieve strong performance in the synthesis but comparatively weaker in sampling. VAE attempts to learn in a continuous latent space, while VQ-VAE \cite{razavi2019generating} and VQ-GAN \cite{esser2021taming} introduce a codebook and learn in a discretized latent space. VQ-VAE and VQ-GAN still achieve strong performance in estimating data distribution. The lower computational cost may allow them to work directly on high-resolution images. Actually, a feasible training requires for a high compression rate of the original images, which is a limitation for generation results. 

\subsection{Unet and Swin-transformer}

Unet is a commonly used architecture in several image processing tasks for its reconstruction capability. It is originally designed for image segmentation in medical area, and widely applied in diffusion model with its encoder-decoder structure. The number and position of skip-connection between encoders and decoders impact its capability to the restoration of images. Although convolutional networks usually act as basic blocks in the Unet, transformer and Swin-transformer have been used as alternative blocks in Unet due to their strong adaptability\cite{chen2021transunet}\cite{fan2022sunet} 

Swin-transformer has been combined with U-net in past works, for image segmentation\cite{cao2023swin} \cite{hatamizadeh2022swin} and restoration\cite{liang2021swinir}. Swin-transformer combines the advantage of transformer and CNN. Compared to convolutional networks, Swin-transformer overcomes the limitation to capture the long-range relation. While compared to a pure transformer structure, hierarchical Swin-transformer decreases the computation with shifted-window attention.

\subsection{Mixture-of-denoising-experts Mechanism}

MOE is widely used but not limited in the NLP domain. In a text-guided diffusion model, denoising step determines which expert to use, and the Unet in different steps act as the experts. During the denoising process for a diffusion model, the aims for all steps are converting random noise into an acceptable image, but the noise ratio of input varies with the time step. When the input data are closer to random noise, the loss of diffusion model drops drastically with the guidance of text. In the opposite, when the input data is close to an image, the loss decreases slower and the model mainly refine details in the images\cite{nichol2021improved}. While considering the observation above, some works attempt to train different models in different time steps with the motivation of MOE.

\section{Dataset}
The base generation model is trained with a subset of 1.5B text-image pairs (Table.\ref{tab:dataset}). 1.32B of them are clean versions of recent publicly available datasets, including Laion-en/zh \cite{schuhmann2022laion}, Conceptual Captions (CC3M) \cite{sharma2018conceptual}, Conceptual 12M (CC12M)\cite{changpinyo2021conceptual}, ImageNet{deng2009imagenet}, Zero Corpus\cite{xie2022zero}, and Wukong\cite{gu2022wukong}. To expand the diversity of concepts, 180M of the data pairs are taken from Common Crawl dataset via the Open Data on Amazon Web Services. To obtain a dataset with high-quality, the above mentioned datasets are further filtered on texts, images, and text-image pair specifically.
In addition, we also process the datasets for evaluation, including the ImageNet Generation benchmark training set and the MS-COCO 2014 validation set.

\begin{table}[h]
  \caption{ Composition of training validation dataset }
\footnotesize
  \label{tab:dataset}
  \begin{tabular}{p{2.3cm}<{\centering}p{1.0cm}<{\centering}p{2cm}<{\centering}p{1.3cm}<{\centering}}
    \toprule
    Dataset  & Original language& Text-image pairs  &Included in training dataset? \\
    \midrule
    \texttt  \   Laion-en & EN & 1,137,886,948  & \XSolidBrush \\
    \texttt  \   Laion-cn & CN & 95,733,575  & \checkmark \\
    \texttt  \   Laion-face & EN & 25,211,628  & \checkmark \\
    \texttt  \      CC12M & EN & 9,984,903  & \checkmark  \\
    \texttt  \   CC3M & EN & 2,712,191  &  \XSolidBrush \\
    \texttt  \   Wukong & CN &  42,770,550 & \checkmark \\
    \texttt  \   Zero-Corps & CN & 16,627,345  & \checkmark\\
    \texttt  \ Common Crawl & CN & 178,200,265   &  \XSolidBrush \\
    \texttt  \ \textbf{Pretrain Total} & & \textbf{1,509,127,405}  &  \textbf{190.328,001}\\
    \texttt  \ ImageNet-train & EN & 1,281,167  &  \checkmark\\
    \texttt  \ COCO 2014-val & EN & 40,504  &  \XSolidBrush\\
    \bottomrule
  \end{tabular}
\end{table}

\subsection{Text Preprocessing}
The text filter is inherited from the coarse filtering of the Massive Data Filtering System (MDFS) in Yuan 1.0, including empty article filtering, bad-words filtering, symbol filtering and converting traditional Chinese to simplified Chinese. Please refer to section 3.1 for more details\cite{wu2021yuan}. Meaningless descriptions, such as serial numbers (1.jpg, image 1, screenshot 1), randomly created names, and “click for more information” are also discarded from texts. The length of image captions should be more than 5 characters. Words that are neither English nor Chinese will also be removed. 
We use the same prompt template for all English captions: 

\textit{This is an image of [CAPTION]}
 
And a Chinese version with the same meaning:

\textit{\begin{CJK}{UTF8}{gbsn}这是一张关于[CAPTION]的图片\end{CJK}}
 
For images with English captions (original text-image pairs from Laion-EN, CC3M, CC12M and ImageNet), the captions are translated into simplified Chinese via OPUS-MT model\cite{tiedemann2020opus}. To select samples with high-quality Chinese descriptions, translated captions are evaluated with gpt2\cite{lagler2013gpt2}. If the perplexity of a caption is large than 6.5, it will be removed. And if the percentage of Chinese characters is less than $70\%$, the translated caption will also be discarded.

The original “caption” for ImageNet are the labels for 1000 classes, and each label is consisted of one or more individual words. First of all, all the labels are manually translated into Chinese. We use as many statement as possible to describe one objective, in order to widen the semantic meaning. For example, the label “syringe” is translated into “\begin{CJK}{UTF8}{gbsn}针筒、针管、注射器\end{CJK}”, which express the similar meaning in Chinese. Secondly, the translated labels are combined with prompt and the serial number of class in Chinese. In this way, the final caption for “syringe” is converted to “\begin{CJK}{UTF8}{gbsn}这是第九百八十四类针筒、针管、注射器的图片\end{CJK}” (This is the nine hundred and eighty-fourth class images of syringe). 

\subsection{Image Preprocessing}
The following preprocessing steps are applied for all datasets and common crawled data:
\begin{itemize}
\item{Remove images with low resolution (less than 64×64).}
\item{Remove images that failed to download.}
\item{Remove duplicated text-image pairs.}
\item{Keep images with aspect ratio lower than 2, to filter out images that are too wide or tall.}
\item{Center-crop all kept images.}
\item{Resize to the images into a uniform resolution (64×64).}
\end{itemize}

Then, we remove images with a higher likelihood of being watermarked or unsafe images via the regarding detectors. To further remove images with illegal content, or meaningless content (QR code, barcode etc), we first collect dozens of examples, then search similar images with their kNN-indices and remove corresponding images.

\subsection{Text-Image Filtering}
To filter out inappropriate text-image pairs, we calculate the similarity of embeddings for images and texts. Multilingual-CLIP \cite{carlsson2022cross}, a multilingual contrastive model with Roberta-Large as text encoder and ViT-B/16+ as vision encoder, is applied to compute the cosine similarity between the text embeddings and image embeddings. Data pairs with cosine similarity less than 0.20 will be trimmed.

\begin{figure}[t]
  \includegraphics[width=\columnwidth]{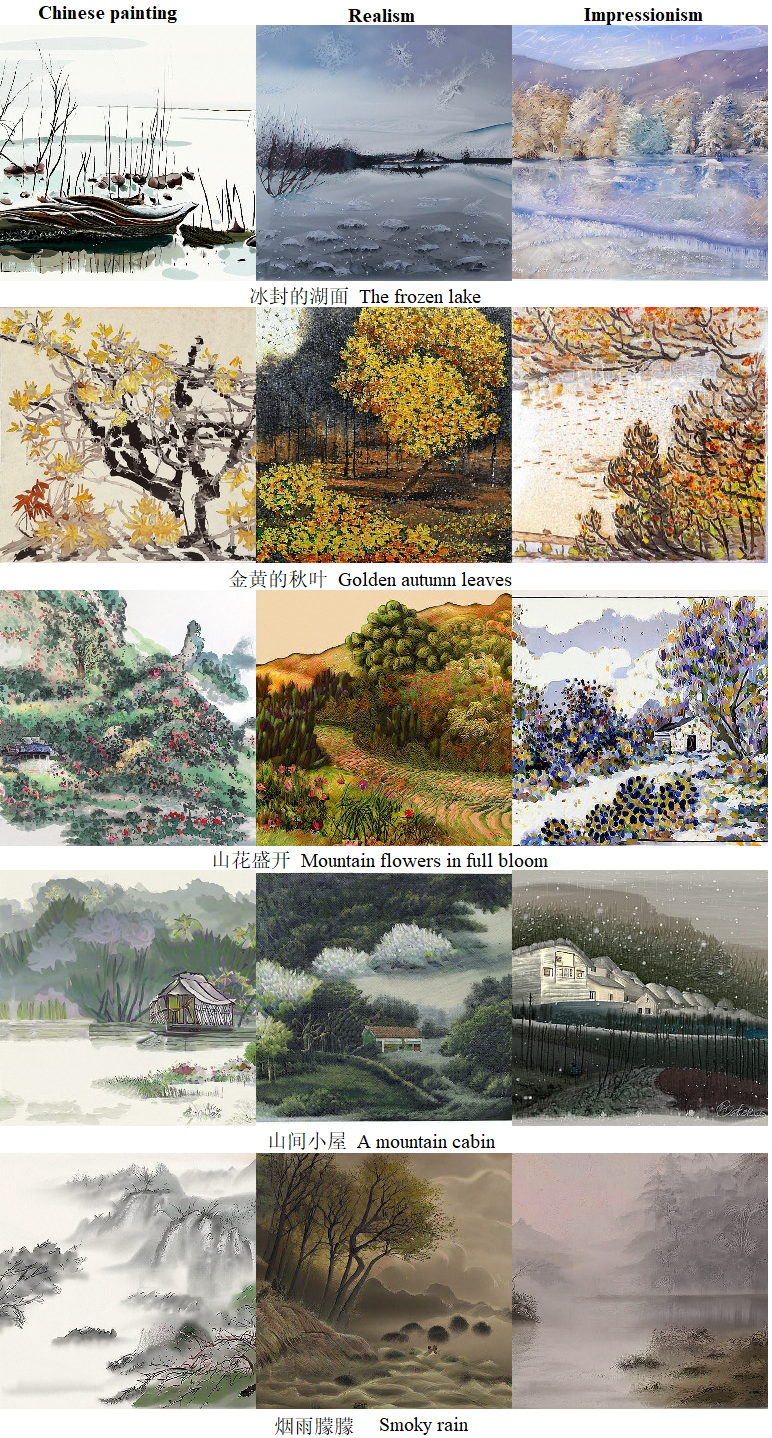}
  \caption{Images in different styles generated with the same caption. (Yuan-TecSwin finetuned on the artworks dataset mentioned in III.D.  }
  \label{fig:style}
\end{figure}

\subsection{Human labeled Dataset}
Because users tend to describe a detailed scene through language, the model's ability to understand the details in the caption and to generate detailed images is essential in content generation. Thus, we also build a prompt artwork dataset with human-written based on the users’ habit. Some of the prompts are collected from user commands submitted to an earlier alpha version of our model. Most of the prompts are contributed by 53 labelers. The labelers are encouraged to describe the details in the painting, including the action, position, color, texture etc. We limit the length of caption from 20 to 150 characters, to balance to quality and the speed. The syntactical structure and manner of speaking are determined by the habits of the labelers, in order to ensure the diversity of the captions. 

The prompted art dataset contains 5043 traditional Chinese paintings and 3651 western paintings.  The final prompt is a random combination of artistic styles, artist, and human-written captions. With the above-mentioned method, we collect 57,995 prompts for Chinese paintings and 34,745 prompts for western paintings.

Considering the clarity and integrity of the paintings, Chinese paintings are selected from Qing Dynasty to modern times, including flower-bird works, mountains-and-water paintings, and genre paintings, etc. If the aspect ratio is smaller than two, the image will be center cropped and resized to the resolution of 64×64. Otherwise, the original image will be cropped into several images according to the short edge, with 50\% contact ratio between the cropped images. Main topic, genre, title and painter are kept as the captions.

Western paintings are divided into 5 classes based on their genres, including realism, impressionism, abstractive painting, illustration, and Japanese paintings. All paintings are center cropped and resized to 64×64. Landscape, architecture, still life, and flowers are the major topics of western paintings. Captions are sentences connected with the information on the main topics, genre, title, and artists, then translated into Chinese with the translation API and manually. Figure \ref{fig:style} shows some generated samples with various painting styles for the same text prompts. In particular, our fine-tuned model captures the distinctive features of different artistic styles and can generate images with artistic value.

\section{Method}
\begin{figure}[t]
  \includegraphics[width=\columnwidth]{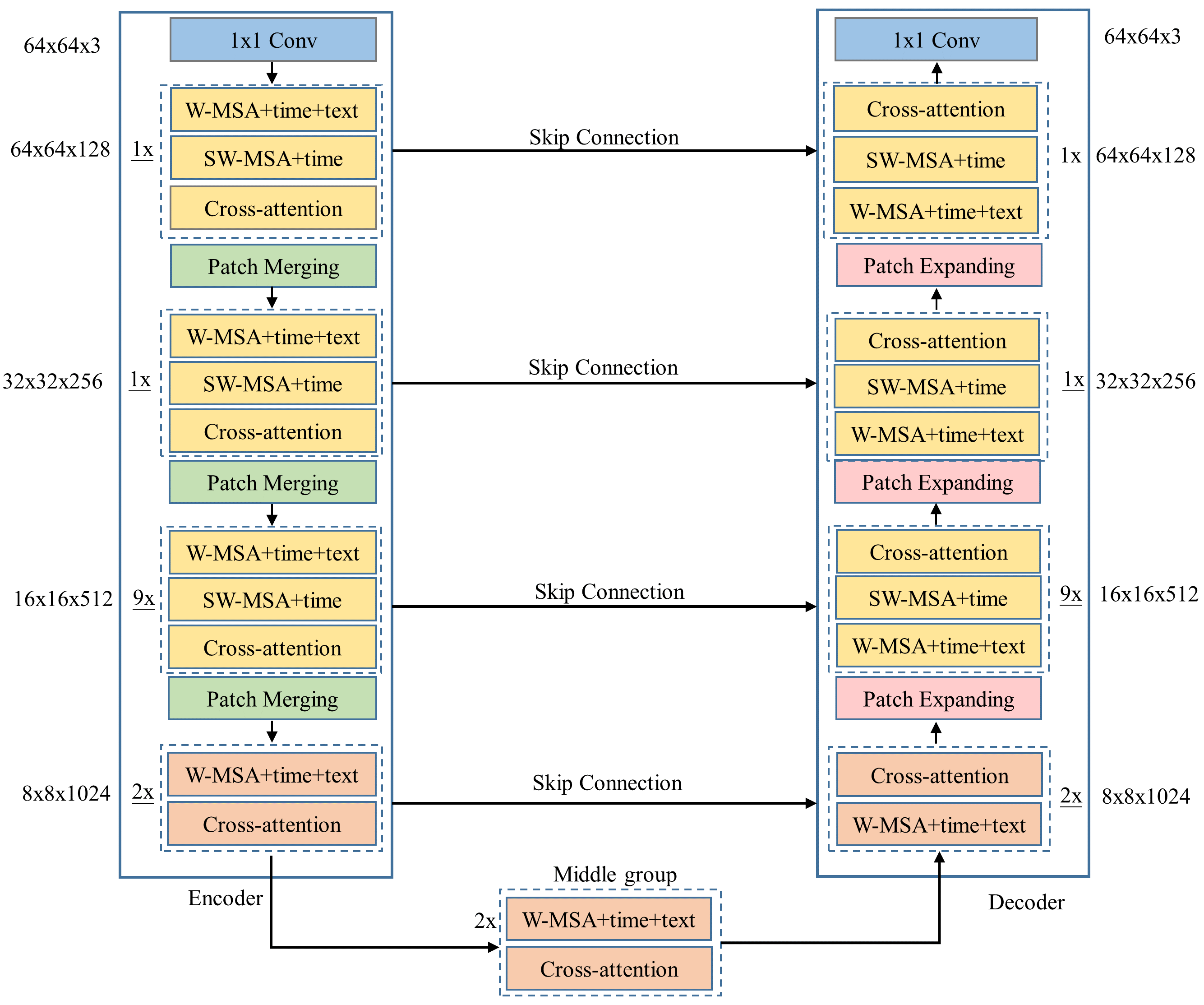}
  \caption{The overall architecture of the Swin-transformer Style base  model} 
  \label{fig:structure}
\end{figure}

 A novel Swin-transformer based diffusion model is designed for our Yuan-TecSwin. The general architecture of the base 64×64 generation model is shown in Figure \ref{fig:structure}, based on the Swin-transformer\cite{liu2021swin} and the Swin-Unet structure\cite{cao2023swin}. We explore the architecture of model in the following three aspects: 1) Structure of Swin-transformer Unet, 2) text conditioning, and 3) Adapted time step searching. Our model is consisted of encoding group, middle group and decoding group.

We present a pipeline of a pre-trained text encoder that maps text to a sequence of embeddings for text guidance, and a Swin based conditional diffusion models that generate images in 64x64 resolution. We use ImageNet generation benchmark for most of the evaluation, comparing the FID for training set. The label combined with the template “This is an image of [label] (\begin{CJK}{UTF8}{gbsn}这是一张关于【label】的图片\end{CJK}) ” are used as input for image generation. 50k images are generated, and the FID are calculated between these generated images and the ImageNet training set. The above captions of ImageNet are translated into Chinese through Baidu Translation API. All images are center cropped and resized to the corresponding resolutions following the approach in Section III. All results displayed in the method section is trained with 10 epoch on ImageNet training set.

\subsection{Text-conditioned Swin-transformer}
We use the Swin-transformer Block instead of traditional CNN blocks in the Unet. The structure of the Swin-transformer block is largely inherited from the Swin-transformer. A window multi-head self-attention (W- MSA) with a LayerNorm (LN) is always followed by a combination of shift-window multi-head self-attention (SW-MSA) and a LN, unless the feature dimension (width/height) of patch equals the window size. Thus, the number of layers in each block is always multiples of 2. A cross-attention layer is applied after each SW-MSA (stage 1-3), and also after W-MSA if the feature dimension equals the window size (stage 4, middle group).

Our model has a symmetric structure about the middle group, with 4 stages in the encoding group and 4 stages in the decoding group. Each stage is consisted of Swin-transformer blocks and a patch merging or a patch expanding layer. The number of Swin Transformer blocks in encoding group is [2,2,18,2], which is the same with that of decoding group. The depth of the model only expands in the third stage, similar to that of Swin-B and ResNet101\cite{he2016deep}. The channel number of the hidden layer in the first stage is set to 128, and the query dimension of each layer is set to 32. The expansion layer of each MLP in SW-MSA and W-MSA is set as 4, and that of each MLP of cross-attention is set as 2 for better performance (Table \ref{tab:mlpratio}).

\begin{table}[h]
\centering
  \caption{ Impact of the mlp-ratio.   }
\footnotesize
  \label{tab:mlpratio}
  \begin{tabular}{p{7cm}<{\centering}p{1cm}<{\centering}}
    \toprule
      Method  &FID \\
    \midrule
    \texttt  \  mlp-ratio for self-attention=2, for cross-attention=2 & 28.82  \\
    \texttt  \  mlp-ratio for self-attention=4, for cross-attention=2 & 26.03\\
    \texttt  \  mlp-ratio for self-attention=4, for cross-attention=4 & 32.35\\
    \bottomrule
  \end{tabular}
\end{table}

The Swin-transformer uses a patch partition to split the image into 16 non-overlapping patches, and the image resolution is 1/16 of its original size. As the resolution of input images for base model is only 64×64, we remove the patch partition layer and apply a 1×1 convolutional layer to project the feature dimension into an arbitrary one. In the encoding group, through each patch merging layer between stages, the output resolution will become 1/2×1/2 of its original size, and the channel will double. While in the decoding group, a patch expanding layer plays a similar role on contrary. In the end, a 1×1 convolutional layer restore the feature map to its original resolution 64×64×3. The total parameter of the model is 341MB.

\begin{table}[h]
  \caption{ Comparison of different down-sample methods.    }
\footnotesize
  \label{tab:updown} 
  \begin{tabular}{p{2cm}<{\centering}|p{4cm}<{\centering}p{1.3cm}<{\centering}}
    \toprule
     &Method  & FID  \\
    \midrule
    \texttt  \   \multirow{3}{*}{Downsample} & Conv2D+Norm  & 26.03  \\
    \texttt  \                               & PatchMerging  & 44.16  \\
    \texttt  \                               & Norm+ Conv2D & 32.48  \\
    \cmidrule(lr){1-3}
    \texttt  \   \multirow{3}{*}{Upsample}  &PixelShuffleUpsample+Norm & 26.03  \\
    \texttt  \                              &PatchExpand & 33.15  \\
    \texttt  \                              &Norm+ PixelShuffleUpsample & 33.12  \\
    \bottomrule
  \end{tabular}
\end{table}

\subsubsection{Downsample and Upsample}
We make an improvement on the downsampling and upsampling modules compared to the structure in Swin-transformer (Table \ref{tab:updown}). Because our model has different feature map scales in different stages, the resizing layers are an important function for the performance. PatchMerging layers down-sample each group of 2x2 neighboring patches with a convolutional layer followed by a LayerNorm, and converting the feature dimension to twice of the original one (rearrange$\xrightarrow{}$ conv2D 1x1 $\xrightarrow{}$rearrange$\xrightarrow{}$ LN).

The structure of PatchExpanding layer is improved based on Dalle-2 \cite{ramesh2022hierarchical}. First of all, a convolutional layer followed by SiLU doubles the input channel and keeps the input resolution. Then, a PixelShuffle function doubles the input resolution and reduce the channel to a quarter. The PatchExpanding layer also ends with a LN (rearrange $\xrightarrow{}$conv2D 1*1 $\xrightarrow{}$SiLU $\xrightarrow{}$ PixelShuffle $\xrightarrow{}$rearrange LN)

We attempted to build an UpSampleCombiner after a patch expanding layer, which appended upsampling results together. We also paid more attention to the final stage of the decoding group. An addition 'final' block, which could be a convolutional layer or a Swin transformer layer, was added after the LN in the last upsampling stage and before the final 1 x 1 convolutional block which restores the image dimension. If there was any UpSampleCombiner, A final convolutional layer was added between an UpsampleCombiner and the final block. However, neither the UpSampleCombiner or the final block or their combination benefited the performance of our model.  

\subsubsection{Swin-transformer Block}
Swin-transformer is applied as the backbone of the diffusion model. Each Swin-transformer block is composed of a W-MSA or SW-MSA, followed by a 2-layer MLP and GeLU. There is always a LayerNorm before each MSA and MLP layer. In the first three stages of the downsampling group and the last three stages of the upsampling group, a SW-MSA follows a W-MSA. While the last stage of the down-sampling group, the first stages of the up-sampling group, and the middle group only have W-MSA groups (Algorithm \ref{alg:alorithm1}).

If a LayerNorm is placed after MSA and MLP layers (res-post norm), and a scaled cosine attention takes the place of a dot product of qk (cosine-attn), the structure within a block will be similar to Swin Transformer V2 \cite{liu2022swin}. A block similar to V2 brings a negative result to our model. A plausible reason is that these are not suitable for a model with only 341M parameters. Res-post-norm and cosine-attn are designed to stabilize the output of each layer, which is effective to model scalability. The log-spaced continuous relative position bias is not experimented with because it is adapted for diverse window resolutions, and we only need 64x64 images as input. 

In most diffusion models, a global self-attention usually exists in the convolutional \cite{saharia2022photorealistic}\cite{nichol2021improved} or transformer blocks\cite{rombach2022high}. As global self-attention improves computation cost by a large margin, no self-attention is applied for the Swin-transformer to achieve a balance between computational complexity and performance benefits. 

\subsubsection{Context Conditioned Swin-transformer (TeC-Swin)}
Diffusion model is trained to revert the diffusion process in each time step with the condition of texts. Context condition (text and time) plays a vital role while converting complete noise into a significant image. 

The base model is trained conditioned with text information at each time step. Texts are encoded into a sequence of tokens, then fed into a multilingual CLIP model for text embeddings. The sequence length for text is set to 512, which is large enough for almost all captions. We use the average of the first, 23rd and 24th layer of token embeddings instead of the last layer. The average text embeddings are separately projected into a certain dimension (batch size*256*512), then concatenated to time (batch size*(256+2)*512). $20\%$ of the input texts are masked. We find that a lower mask percentage leads to better performance. In our experiment, the text condition can decrease the FID by 23.5\%.

To increase the interaction between the image and the corresponding time and text information, the structure of the Swin-transformer is further adjusted in 1. position of scale shift, 2. concatenating context with W-MSA in key and value dimension, 3. position of cross attention, and 4. percentage of text mask.

The embedding of context is conditioned with an embedding pooled vector and added to diffusion time-step embedding. We make thorough experiments on the position of scale- shift, and find that the position of scale-shift influence the performance of model by large. The original position of scale-shift is after each Swin-transformer Block, individual of residual model. When we put the scale-shift inside the residual module, it increases the FID by nearly 15 points. Table \ref{tab:scale-shift} displays our exploration on the position and related accuracy.

\begin{table*}[h]
\centering
  \caption{ Impact of the scale-shift position.   }
\footnotesize
  \label{tab:scale-shift} 
  \begin{tabular}{p{2cm}<{\centering}p{10cm}<{\centering}p{3cm}<{\centering}}
    \toprule
    id &Position of scale-shift & FID  \\
    \midrule
    \texttt  \   1 & after Swin, individual of residual  & 47.24  \\
    \texttt  \   2 & Inside residual, short-cut$\xrightarrow{}$scale-shift$\xrightarrow{}$GeLU$\xrightarrow{}$Norm$\xrightarrow{}$WA & 31.97  \\
    \texttt  \   3 & Inside residual, short-cut$\xrightarrow{}$GeLU$\xrightarrow{}$Norm$\xrightarrow{}$WA & 33.96  \\
    \texttt  \   4 & Inside residual, short-cut$\xrightarrow{}$Norm$\xrightarrow{}$scale-shift$\xrightarrow{}$GeLU$\xrightarrow{}$WA & 28.82  \\
    \texttt  \   5 & Inside residual, short-cut$\xrightarrow{}$Norm$\xrightarrow{}$scale-shift$\xrightarrow{}$WA & 29.10  \\
    \texttt  \   6 & Inside residual, short-cut$\xrightarrow{}$Norm$\xrightarrow{}$scale-shift$\xrightarrow{}$GeLU$\xrightarrow{}$Norm$\xrightarrow{}$WA & 37.74  \\
    \texttt  \   7 & after Swin block, individual of short-cut  & 77.63  \\
    \texttt  \   8 & Replace the LayerNorm with Norm+scale-shift+GeLU in each Swin Block  & 33.06  \\
    \texttt  \   9 & Replace the LayerNorm with Norm+scale-shift+GeLU in each Swin Block  & 32.31  \\
    \texttt  \   10 & Inside residual, short-cut$\xrightarrow{}$Norm$\xrightarrow{}$scale-shift$\xrightarrow{}$SiLU$\xrightarrow{}$WA & 29.10  \\
    \bottomrule
  \end{tabular}
\end{table*}

A key design element lies in the concatenation of context with W-MSA in kv (key and value), which bridges the context to image features in every self-attention computed within the window. Due to the difference of context dimension and image feature dimension, the concatenation is facilitated with a middle variable attn. First, we calculate the self-attention (attn) of the image by multiplying the query and key. The dimension of context is converted from (batchsize*258*512) to (b, 258, num-head*head-dim), and an attn-c (attention with context) of context is calculated by multiplying query of images and key of context. Then, we concatenate attn with attn-c and concatenate v of the image feature with v-c. The multiplication of combined attn and combined v is the output of window attention.

Cross-attention layers attending over context embeddings are added in each stage after SW-MSA. When the width and heights equal the window size, there will be no SW-MSA in the stage, then the cross-attention layer will follow each W-MSA. Every cross-attention layer is also followed by an MLP layer with expansion layer as 2, which is the same as W-MSA. Both the position and the amount of cross-attention influence the model performance considerably. An additional cross-attention after every W-MSA brings no further improvement.

\begin{algorithm}[H]
\caption{Pesudo code implementation for the position of scale-shift}\label{alg:alorithm1}
\begin{algorithmic}
\STATE 
\STATE  short-cut=x
\STATE  x=norm(x)
\STATE  x=x*(scale+1)+shift
\STATE  x=Gelu(x)
\STATE  x=short-cut + attn(x)
\end{algorithmic}
\label{alg1}
\end{algorithm}

\subsection{Text encoder and text embeddings}For a text-conditioned diffusion model, a text encoder that achieves accurate understanding of the text prompts is necessary to guide image generation. In general, there are two types of text encoder that can be chosen for diffusion models. The first kind is represented by CLIP \cite{pmlr-v139-radford21a}, which trains a text encoder and an image encoder with contrastive learning strategy. CLIP is generally considered to be good at capturing the correspondence of cross-modal representations. Since the datasets used by CLIP and the text-to-image generation task are both text-image pair, it makes the two tasks especially domain-consistent. The other is represented by the large language model \cite{2020Exploring} trained only on the text corpus. As there is no restriction that text must be paired with an image, the scale of the corpus can be an order of magnitude larger than the text-image pair, which makes it possible to train larger language models. Text encoders with more parameters are usually better at understanding complex text.

Some experiments are carried out in M-CLIP\cite{carlsson2022cross} and T5-XXL, to explore the pros and cons of two types of text encoders in our work. We choose a pre-trained M-CLIP model as the text encoder for the following reasons. 1. T5-XXL displays a similar result as  CLIP on automatic evaluation benchmarks (e.g. FID 30k on MS-COCO) with much more parameters (11B v.s. 63M). It is not computationally efficient to introduce a giant model during training without significant performance improvements. 2. Text prompts in text-to-image generation tasks are rarely too complex or too long. It does not require much in terms of the ability of large language models to capture the complexity of text. However, the domain-consistency of image-text pairs between CLIP and diffusion models is dominant. 3. To support text-to-image generation in Chinese, we select a pre-trained M-CLIP as text encoder. Specifically, we introduce a pre-trained M-CLIP with XLM-RoBERTa Large model as our text encoder.   

We further perform a thorough research on the ability of the text encoder, by comparing the output of different layers of the text encoder. We observed that images created using the average of the first, 22nd, and the last layer as text embeddings (FID = 26.03) perform better than only using the average of the first and final layer (FID = 29.03), or the output of last layer (FID = 31.23).

\subsection{Adapted time step searching}
During inference, the number of diffusion timesteps greatly influences the performance of the model, as the different values determine the best quality of the sample. With ImageNet Generation Benchmark, we searched through time steps and found that 190 is the most appropriate value (Figure \ref{fig:condscale}). There is barely any certain trend between the timestep and the accuracy, but we can always get a minimum via experiments.

\begin{table}[h]
\centering
  \caption{Comparison between MOE and adapted time-step searching  }
\footnotesize
  \label{tab:MOE}
  \begin{tabular}{p{2cm}<{\centering}p{1cm}<{\centering}p{3cm}<{\centering}p{1.2cm}<{\centering}}
    \toprule
      & num of model &FID &improvement  \\
    \midrule
    \texttt  \    Yuan-TecSwin & 1 & 1.56$\xrightarrow{}$1.37 (imagenet)  & 12.2\%  \\
    \texttt  \    Ernie-ViLG 2.0  & 2 &  6.75(MS-COCO) & \\
    \texttt  \    e-diff  & 3 & 7.35$\xrightarrow{}$6.95(MS-COCO) & 5.5\% \\
    \bottomrule
  \end{tabular}
\end{table}

Inspired by the idea of a mixture of denoising experts, we uniformly divide the denoising steps into several stages to refine the inference process. We further employ different steps and apply the same model in each stage, instead of training different experts. Specifically, we divide the 190 timesteps uniformly into 19 stages, thus each stage contains 10 substeps. The substep in each stage is further searched individually from stage 0 to stage 18, in order to get its local optimized number of substep. In stage 0, the input data are close to total noise, and in stage 18, the input data are almost a meaningful image. This method ensures a finer sampling at the first and final several stages and coarser sampling in the middle. With the adapted time step searching, we improve the FID from 1.565 to 1.372 (Table \ref{tab:MOE}).

\subsection{Training Details}
We train the model with a global batch size of 1024. The model is optimized by Adam, with a cosine learning rate of 0.00015, min learning rate=0.000015, learning rate warm-up fraction=0.005, and weight decay=0.

We further fine-tune Yuan-TecSwin with hunan-labeled artworks and traine the dataset for 5 epochs with a constant learning rate (1*10-6). 5 epochs ensure a balance between command understanding and generation diversity in our work.

\section{Experiments}
We evaluate our method on ImageNet generation benchmark in the resolution of 64×64. The training set of ImageNet consists of approximately 1.2M images, and the validation set has 50k images. ImageNet contains 1000 categories in total. Following previous works, for class-conditional image generation on ImageNet, we train our model on the whole training set, and generate 50k samples to report the FID against the training set. In addition to the ImageNet, we also compare results on MS-COCO validation set, for the comparison between CNN-based and Swin-based model. We additionally evaluate our model on MS-COCO 2014 with zero-shot FID-30k, where 30k captions from validation set are randomly selected to generate samples.

\subsection{Augmentation and Evaluation on ImageNet }
Using the ImageNet Generation Benchmark, we compare our model with CDM\cite{ho2020denoising}, ADM, improved DDPM and BigGAN, for 64*64 resolution on training set (Tabel \ref{tab:11}). Yuan 2.0 outperforms the CDM score by almost 0.1 in terms of FID, with only one 64*64 diffusion model. CDM builds a two-stage cascading pipeline with a 32*32 base model and a 32*32 to 64*64 super-resolution model. 

\begin{table}[h]
  \caption{ Comparison of different model on ImageNet Generation Benchmark }
\footnotesize
  \label{tab:11}
  \begin{tabular}{p{5cm}<{\centering}p{2cm}<{\centering}}
    \toprule
     Model & FID  \\
    \midrule
    \texttt  \   Yuan-TecSwin  & 1.37 \\
    \texttt  \   CDM  & 1.48 \\
    \texttt  \   ADM  & 2.07 \\
    \texttt  \   Improved DDPM  & 2.92 \\
    \texttt  \   BigGAN-deep  & 4.06 \\
    \bottomrule
  \end{tabular}
\end{table}

\begin{figure}[t]
\centering
  \includegraphics[width=\columnwidth]{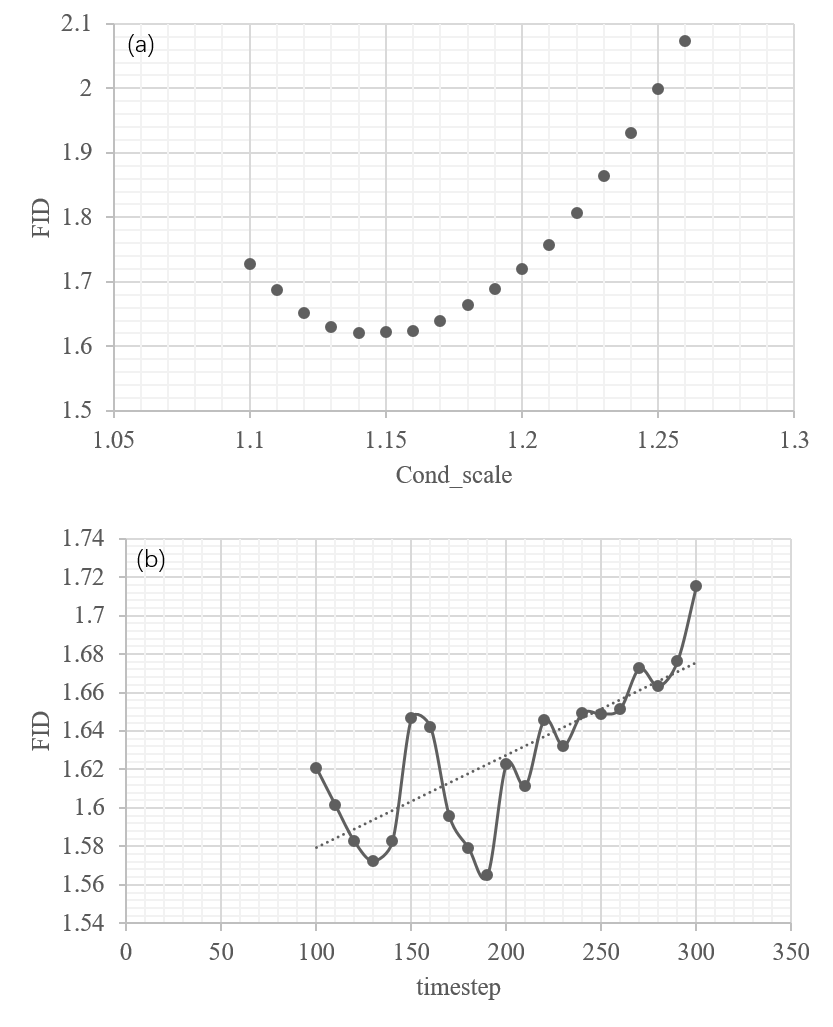}
  \caption{ Impact of (a) cond-scale and (b) timestep on FID. 
} 
  \label{fig:condscale}
\end{figure}

To improve the performance for image generation, we search the hyper-parameter of cond-scale and number of timestep (Figure \ref{fig:condscale}). Cond-scale influences the text guidance weight. It is first searched in a large scale from 1.0 to 7.0 based on previous experience, then refined within 1.1 and 1.26 (Figure \ref{fig:condscale}.a). With the augmentation on guidance, the FID score degrades from 1.72 to 1.62.

The number of time-step is another critical factor that may improve sample quality. With the most proper cond-scale (1.14), we search the value of time-step from 100 to 400. Although there is no quantitative relation between time step and FID, there is a declining trend ever since 200. The searching of time step allows the FID decrease from 1.62 to 1.56. Then followed by adapted time step searching method mentioned in section 4.3, the final FID changes from 1.56 to 1.37.

\subsection{Results and evaluation on MS-COCO}
We compare the performance of CNN-based and Swin-based diffusion model on MS-COCO validation set, with FID-30k score. All results are reported at 64×64 resolution in this section.
The CNN-based model is also a 64×64 text-conditional diffusion model, trained with the same dataset for Swin-based model. The architecture is adapted from the base Unet in GLIDE, and the parameter is no less than 16.3 billion for a better performance. The CNN-based model is also trained via continuous timesteps and cosine noise schedule. In spite of model structure within the Unet, we keeps all other conditions the same. 
As a result, the swin-based diffusion model achieves a zero-shot FID-30k on MS-COCO at 6.201, outperforming the CNN-based model by 0.9 points (7.18). It is worthy to notice that the Swin-based model only has 341M parameters that is one fifth of its rival.

\subsection{Turing test for image generation}
In order to test the generation performance more thoroughly, we arbitrarily selected 23 images created by model as the experimental group, and 23 paintings or photos by human as the control group. The questionnaire is consisted of 7 photorealistic images, 5 traditional Chinese paintings, 7 western paintings, and 4 illustrations. All the 46 images are center cropped and resized to the same resolution 256×256. 

Respondents need to judge whether an image is created by human or by a model.
We collect 124 valid questionnaires in total. The human accuracy at detecting images created by our model is $51.4\%$, which implies that participants are unable to distinguish human-created and model-created images. 
The generated photos ($48.8\%$) indicate excellent capacity in detail description. The generation of artworks indicate that our model is able to create paintings with a certain genre, or a mixing of different genres.

According to our interview, respondents pay more attention on details regarding to photorealistic images, while to paintings and illustrations, they care more for the displayed color tones. This may lead to an extra bias towards artworks. Respondents tend to choose an image with higher brightness as the generated one, and an image with lower brightness as the human-painted one. They regard the grey or yellow tones as “a sign of time”. In fact, our model is able to create either brighter or darker images. Yuan-TecSwin also displays the capacity in describing details.

\subsection{The selection of Guidance scales}
Guidance scale can affect sampling quality significantly.   For the Swin-based model, slightly smaller guidance scales help to achieve a better FID, and too large guidance scales improve the alignment with text prompt but compromise image fidelity. We reach the best FID on ImageNet 64×64 with 1.14.

\section{Conclusion}
Yuan-TecSwwin is the first swin-based text conditioned diffusion model trained in with the largest Chinese multi-modal image-text dataset. It archives the SOTA FID with only 341M parameter. To improve the image generation quality, we make thorough experiments on the structure of swin model, output of text encoder, and the key hyperparameters influencing inference process. Human evaluation implies the difficulty to tell generated paintings by Yuan-TecSwin, and existing masterpieces by human.

{
    \bibliographystyle{IEEEtran} 
    \bibliography{egbib}
}


\newpage

\vfill

\end{document}